%% file: main.tex
\documentclass[10pt,twocolumn,letterpaper]{article}

\usepackage[pagenumbers]{cvpr} %

\usepackage{graphicx}
\usepackage{amsmath}
\usepackage{amssymb}
\usepackage{booktabs}
\usepackage{nicefrac}
\usepackage{lipsum}
\usepackage{color}
\usepackage{xspace}
\makeatletter
\@namedef{ver@fixltx2e.sty}{}
\makeatother
\usepackage{dblfloatfix}

\newcommand\blfootnote[1]{%
  \begingroup
  \renewcommand\thefootnote{}\footnote{#1}%
  \addtocounter{footnote}{-1}%
  \endgroup
}

\usepackage[pagebackref,breaklinks,colorlinks]{hyperref}
\usepackage{xcolor}
\definecolor{myblue}{HTML}{226DA3}
\hypersetup{
    citecolor=myblue
}

\newcommand{\data}{\mbox{\textsc{{Objaverse}}}\xspace}

\newcommand{\datalvis}{\mbox{\textsc{{Objaverse-LVIS}}}\xspace}

\usepackage[capitalize]{cleveref}
\crefname{section}{Sec.}{Secs.}
\Crefname{section}{Section}{Sections}
\Crefname{table}{Table}{Tables}
\crefname{table}{Tab.}{Tabs.}

\def\cvprPaperID{3974} %
\def\confName{CVPR}
\def\confYear{2023}

\begin{document}

\title{Objaverse: A Universe of Annotated 3D Objects}

\author{\\[-0.3in]\textbf{Matt Deitke$^{\dagger\psi}$, Dustin Schwenk$^\dagger$, Jordi Salvador$^\dagger$, Luca Weihs$^\dagger$, Oscar Michel$^\dagger$}\\\textbf{Eli VanderBilt$^\dagger$, Ludwig Schmidt$^\psi$, Kiana Ehsani$^\dagger$, Aniruddha Kembhavi$^{\dagger\psi}$, Ali Farhadi$^\psi$}\\
$^\dagger$PRIOR @ Allen Institute for AI, $^\psi$University of Washington, Seattle\\
\href{https://objaverse.allenai.org/}{\texttt{objaverse.allenai.org}}
}

\newcommand{\w}{1.329in}
\newcommand{\h}{1in}

\twocolumn[{
\renewcommand\twocolumn[1][]{#1}
\maketitle
\vspace*{-0.25in}
\centering
\captionsetup{type=figure}\includegraphics[width=\textwidth]{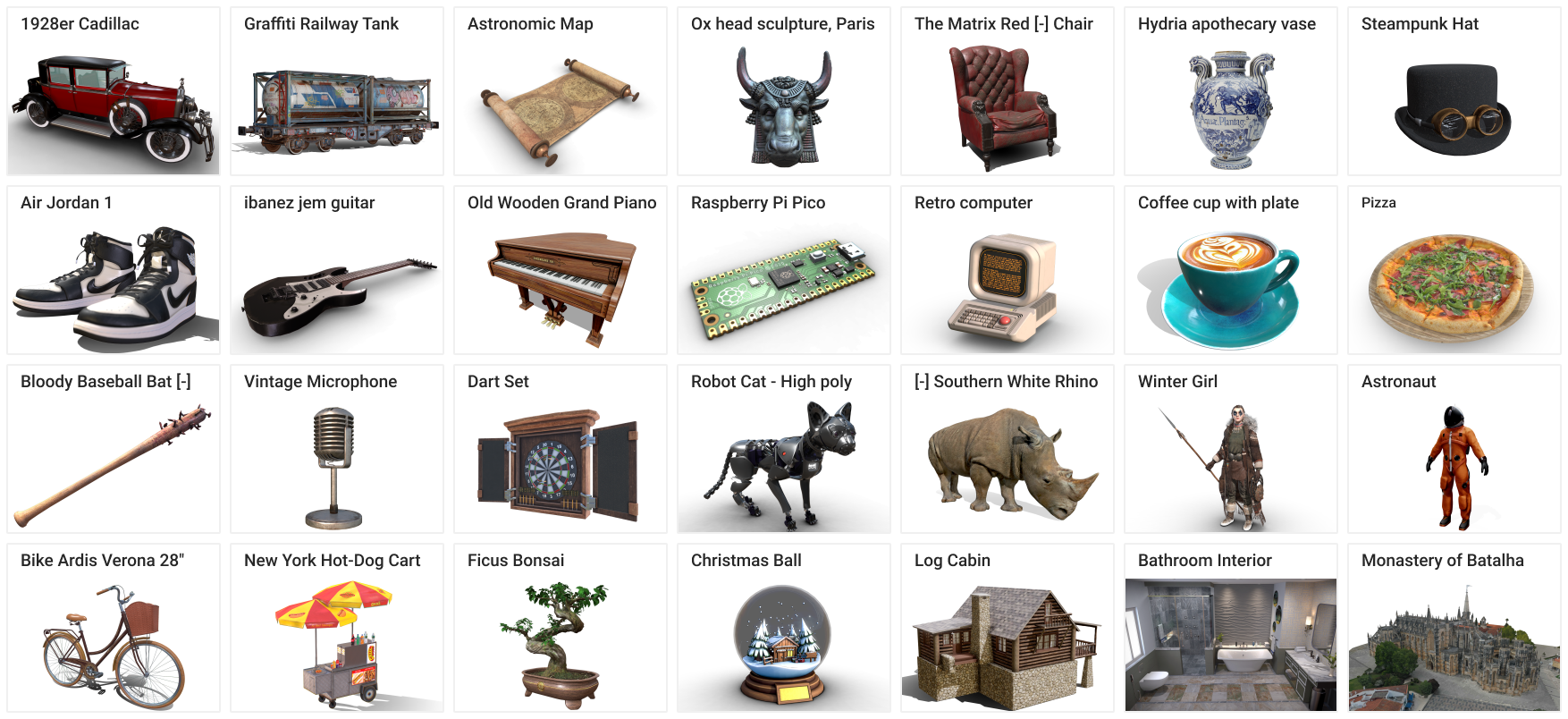}
\captionof{figure}{Example instances from our large-scale 3D asset dataset \data. \data 3D assets are semantically diverse, high-quality, and paired with natural-language descriptions.\\[0.15in]}
\label{fig:teaser}
\vspace{-1em}
}]

\maketitle

\blfootnote{Correspondence to $<$mattd@allenai.org$>$.}

\vspace*{-0.10in}

\begin{abstract}
\vspace{-0.125in}
Massive data corpora like WebText, Wikipedia, Conceptual Captions, WebImageText, and LAION have propelled recent dramatic progress in AI. Large neural models trained on such datasets produce impressive results and top many of today's benchmarks. A notable omission within this family of large-scale datasets is 3D data. Despite considerable interest and potential applications in 3D vision, datasets of high-fidelity 3D models continue to be mid-sized with limited diversity of object categories. Addressing this gap, we present Objaverse 1.0, a large dataset of objects with 800K+ (and growing) 3D models with descriptive captions, tags, and animations. Objaverse improves upon present day 3D repositories in terms of scale, number of categories, and in the visual diversity of instances within a category. We demonstrate the large potential of Objaverse via four diverse applications: training generative 3D models, improving tail category segmentation on the LVIS benchmark, training open-vocabulary object-navigation models for Embodied AI, and creating a new benchmark for robustness analysis of vision models. Objaverse can open new directions for research and enable new applications across the field of AI.

\end{abstract}

\vspace{-0.25in}
\section{Introduction}
\label{sec:intro}

\newcommand{\s}{0.05in}

\input{01_intro_new.tex}

\vspace{-0.05in}

\section{Related Work}
\label{sec:related-work}

\textbf{Large scale datasets.} 
Scaling the size and scope of training datasets has widely been demonstrated to be an effective avenue of improvement for model performance. In computer vision, the adoption of early large scale datasets such as Imagenet\cite{russakovsky2015imagenet,deng2009imagenet} and MS-COCO\cite{lin2014microsoft} has dramatically accelerated progress on a variety of tasks including classification, object detection, captioning, and more. Ever since, the diversity and scale of datasets have continued to grow. YFCC100M is a dataset of 99.2M images and 800K videos\cite{thomee2016yfcc100m}. OpenImages\cite{kuznetsova2020open} is a large scale dataset of 9M images that contains labeled subsets bounding boxes, visual relationships, segmentation masks, localized narratives, and categorical annotations. Massive web-scraped datasets containing image-text pairs such as Conceptual Captions\cite{sharma2018conceptual}, WIT\cite{srinivasan2021wit}, and LAION\cite{schuhmann2021laion,schuhmann2022laion} have seen increased popularity recently as they have been used to train impressive models for vision-language representation learning\cite{radford2021learning,ilharco_gabriel_2021_5143773,jia2021scaling}, text-to-image generation\cite{ramesh2021zero,ramesh2022hierarchical,rombach2022high,jia2021scaling}, and vision-language multitasking\cite{cho2021unifying,tan2019lxmert,wang2022image,chen2022pali}.

\textbf{3D datasets.}
Current large-scale 2D image datasets offer three crucial components that benefit learning: scale, diversity, and realism. Ideally, models that reason about 3D objects should have access to datasets that meet these same criteria. However, of the numerous 3D object datasets that currently exist, none are able to excel in all three categories to the same degree as their 2D counterparts. 
Datasets such as KIT\cite{kasper2012kit}, YCB\cite{calli2015benchmarking}, BigBIRD\cite{Singh2014BigBIRDAL}, IKEA\cite{lim2013parsing}, and Pix3D\cite{sun2018pix3d} provide image-calibrated models over a diverse set of household objects, but severely lack in scale with only a few hundred objects at most. EGAD\cite{morrison2020egad} procedurally generates 2K objects for grasping, but produces objects that are not that realistic or diverse. Slightly larger datasets of photo-realistic objects include GSO\cite{downs2022google}, PhotoShape\cite{park2018photoshape}, ABO\cite{collins2022abo} and 3D-Future\cite{fu20213d}, and ShapeNet\cite{chang2015shapenet} with object counts in the tens of thousands, see Fig.~\ref{fig:shapenet-comparison} for comparisons between \data and these datasets. Datasets for CAD models, such as ModelNet\cite{wu20153d} and DeepCAD\cite{wu2021deepcad}, and ABC\cite{koch2019abc} do not include textures or materials, which limits their ability to represent objects that could plausibly be found in the real world. Datasets of scanned 3D objects and environments are valuable for real-world understanding\cite{couprie2013indoor,dai2017scannet,choi2016large,levoy2000digital}, but are quite small and limited. In addition to containing numerous artist designed objects, \data contains many scanned assets, making it a useful source of data for learning from real-world distributions.

While rapid progress has been made in developing datasets that combine image and text, in contrast, only a few datasets that pair language and 3D data exist. Text2Shape\cite{chen2018text2shape} released a dataset of 15,038 chairs and tables from ShapeNet each with around 5 text captions, giving 75,344 total text-shape pairs. ShapeGlot\cite{achlioptas2019shapeglot} released the CiC (Chairs in Context) dataset which contains 4,511 chairs from ShapeNet along with 78,789 descriptive utterances generated from a referential game. Due to the small scale and limited diversity of these datasets, current SoTA text-to-3D models\cite{michel2022text2mesh,poole2022dreamfusion,jain2022zero} forgo the use of 3D datasets entirely and instead rely on 2D image-text supervision.

\vspace{-0.075in}

\section{Objaverse}
\label{sec:objaverse}

\data is a massive annotated 3D dataset that can be used to enable research in a wide range of areas across computer vision.
The objects are sourced from Sketchfab, an online 3D marketplace where users can upload and share models for both free and commercial use. Objects selected for \data have a distributable Creative Commons license and were obtained using Sketchfab’s public API. Aside from licensing consideration, models marked as restricted due to objectionable or adult thematic content were excluded from the dataset.

\begin{figure*}[t!]
     \centering
          \begin{subfigure}[b]{0.48\textwidth}
         \centering
         \includegraphics[width=\textwidth]{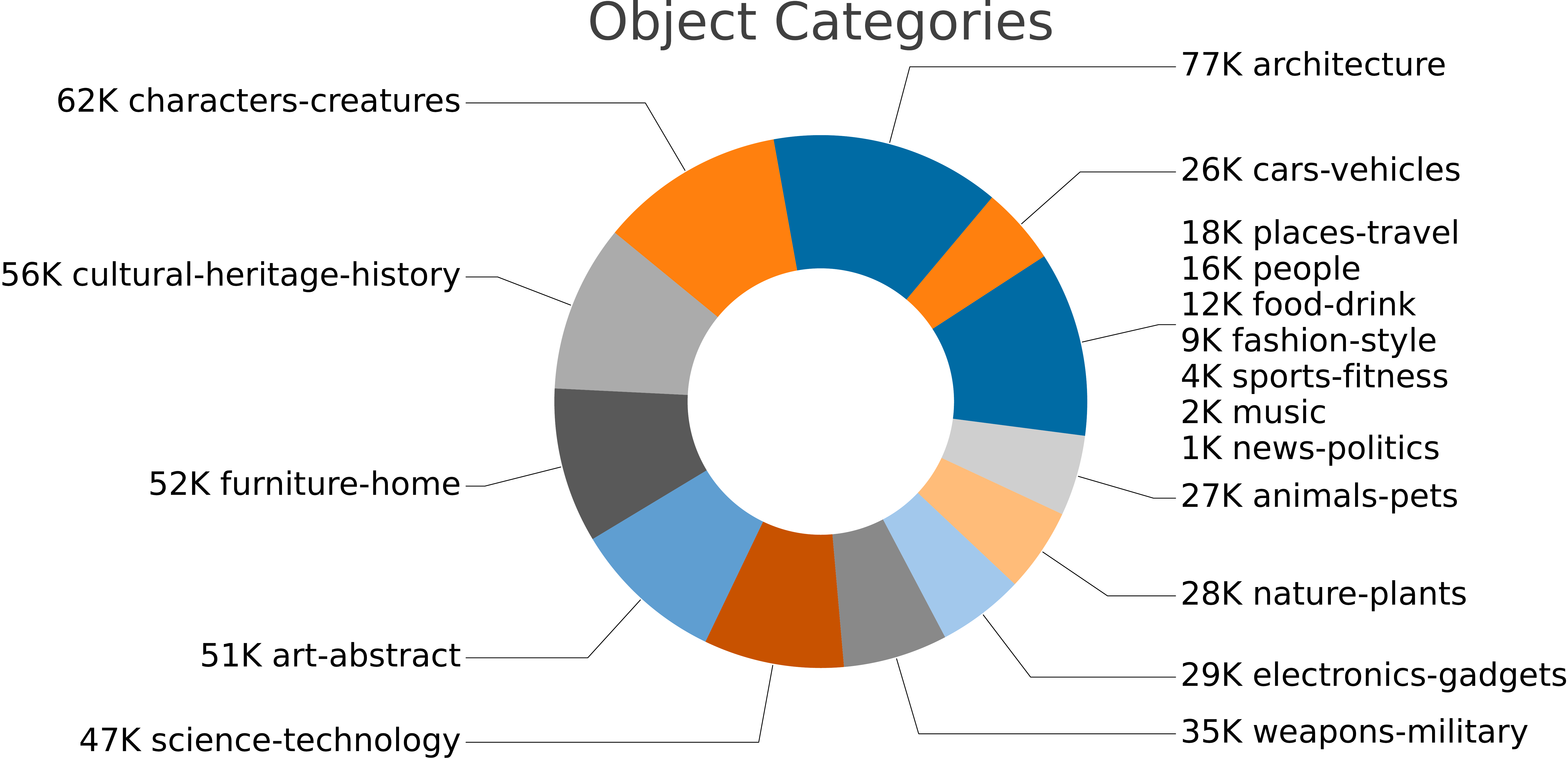}
         \caption{Breakdown of high-level \data categories}
         \label{figures:category_donut} 
     \end{subfigure}
     \hfill
     \begin{subfigure}[b]{0.48\textwidth}
         \centering
         \includegraphics[width=\textwidth]{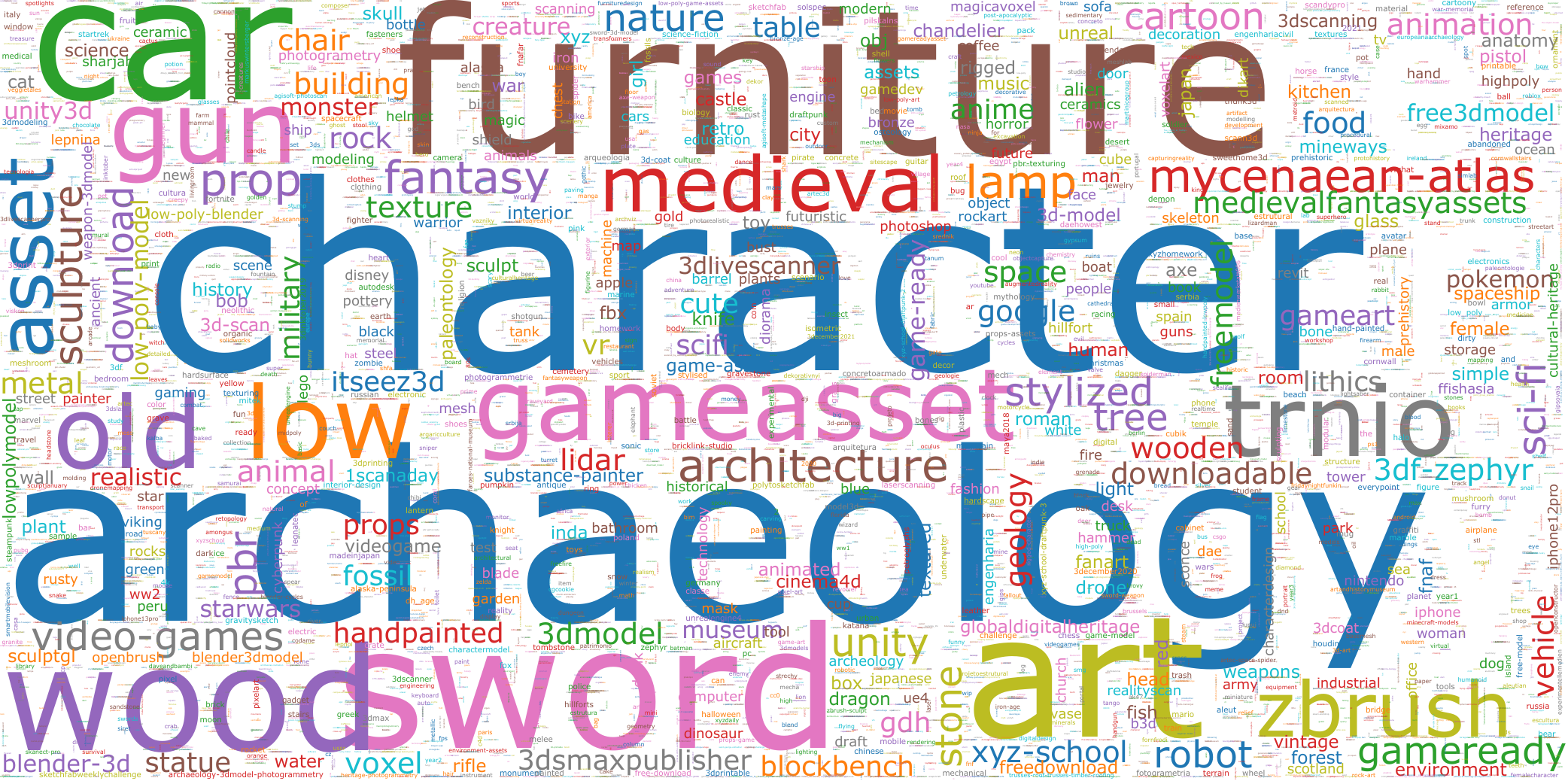}
         \caption{Word cloud of \data metadata tags.}
         \label{figures:tag_cloud} 
     \end{subfigure}
     \\[0.15in]
     \begin{subfigure}[b]{0.48\textwidth}
         \centering
         \includegraphics[width=\textwidth]{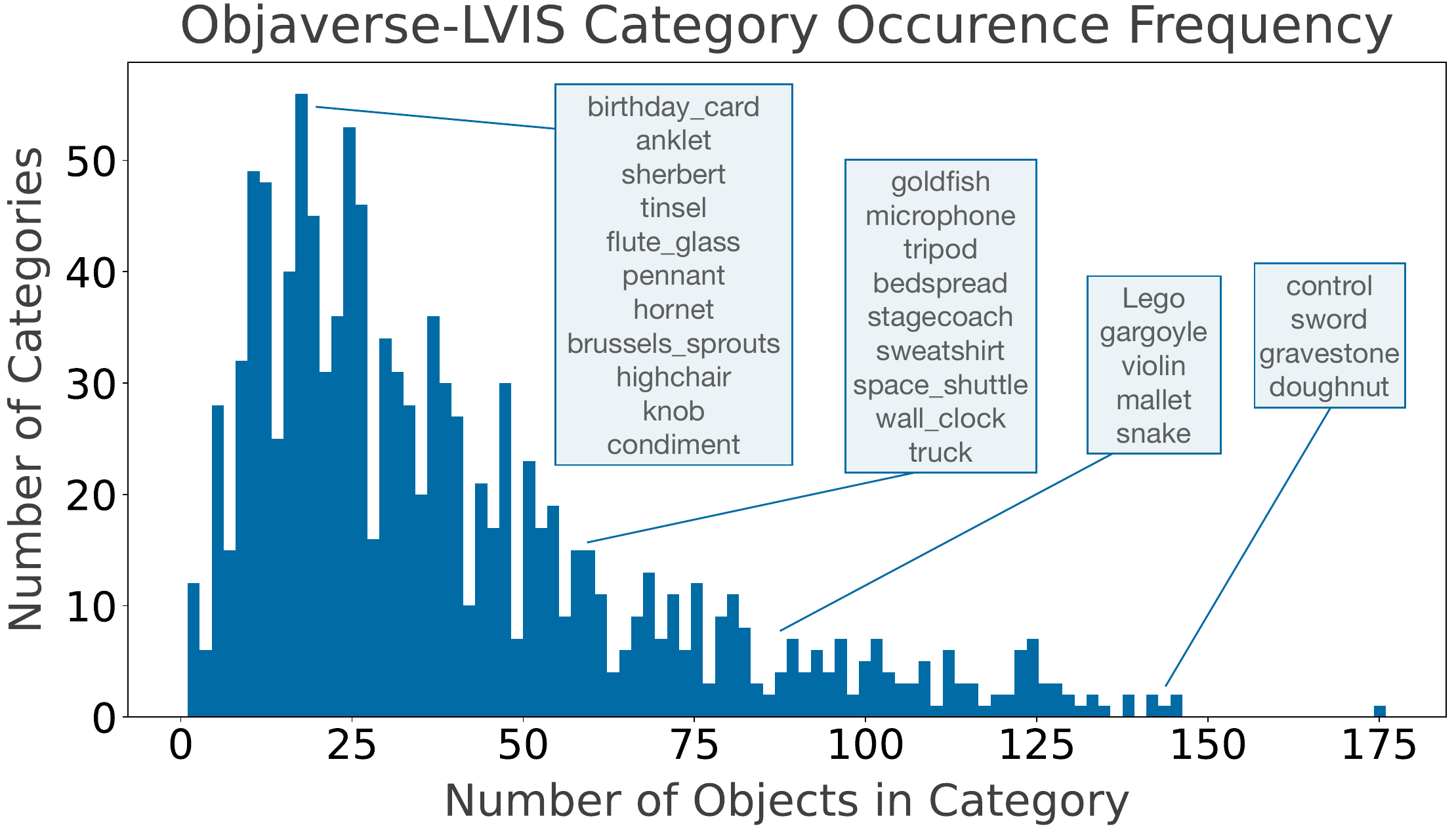}
         \caption{\datalvis category occurrence distribution.}
         \label{figures:lvis_cat_dist} 
     \end{subfigure}
     \hfill
     \begin{subfigure}[b]{0.495\textwidth}
         \centering
         \includegraphics[width=\textwidth]{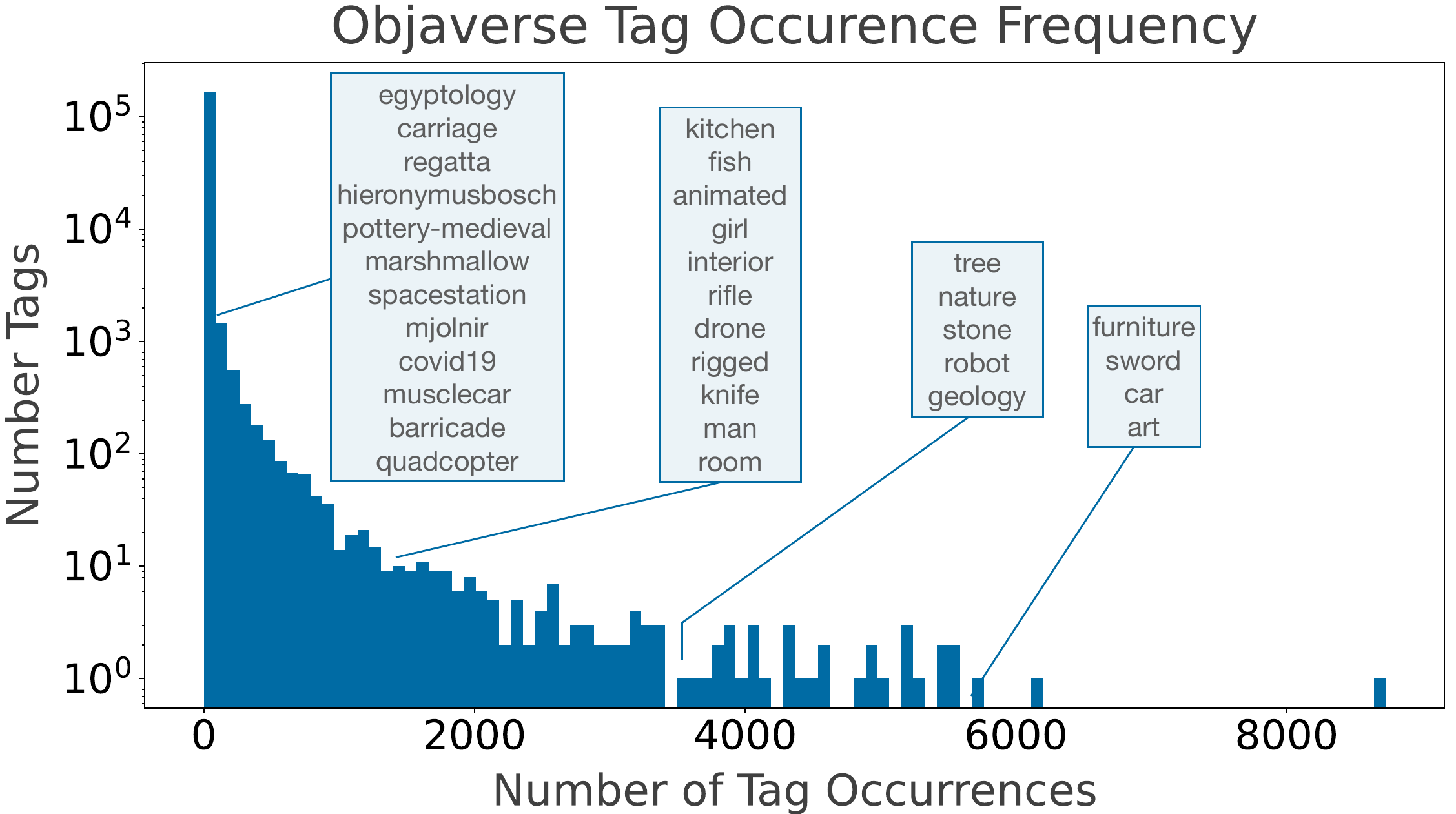}
         \caption{\data tag occurrence distribution.}
         \label{figures:tag_dist} 
     \end{subfigure}
     \caption{\textbf{\data statistics.} (a) All 18 high-level categories present in \data's metadata with their corresponding number of occurrences. The relative share of most popular categories are evenly split, with a small number of less frequently categories. (b) A sample of several thousand popular object tags found in \data log-scaled by their frequency. (c) A histogram of fine-grained \datalvis categories with representative members from several bins highlighted.
     (d) A histogram of \data tags with representative members from several bins highlighted (note y-axis log scale). Tags from the low-occurrence side of the distribution correspond to unique objects that, taken individually, are rarely seen in the world. Frequently used tags like "furniture" and "car" reflect their real-world normalcy, but the high frequency of assets like "sword" diverge from their real-world counterparts.} 
     \label{fig:metadata vis}
     \vspace{-1em}
\end{figure*}

\begin{figure*}[t!]
    \centering
    \includegraphics[width=1\textwidth]{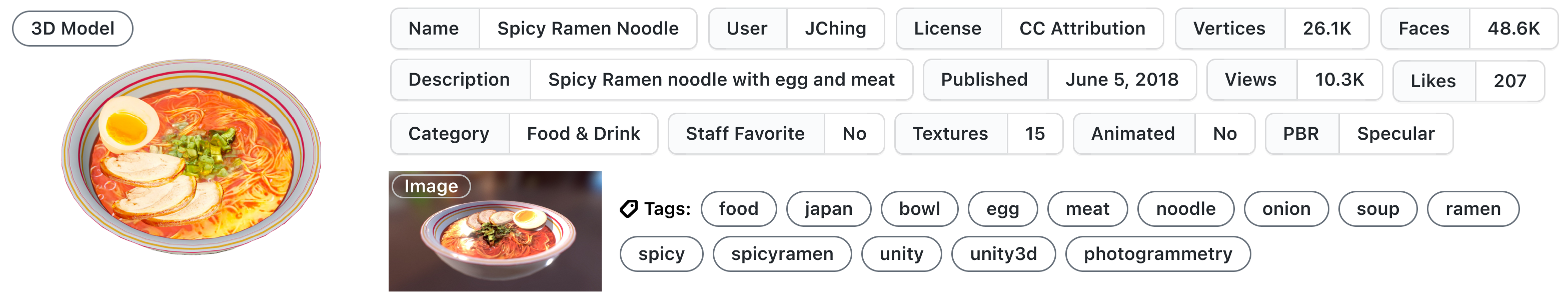}
    \caption{An example of metadata available for each object in \data. Each uploaded object has a 3D model, user-selected rendered thumbnail image, name, description, tags, category, and stats, among additional metadata.}
    \label{fig:sf-metadata}
    \vspace{-1em}
\end{figure*}

\textbf{Model metadata.}
\data objects inherit a set of foundational annotations supplied by their creator when uploaded to Sketchfab. Figure~\ref{fig:sf-metadata} shows an example of the metadata available for each model. The metadata includes a name, assignments to a set of fixed categories, a set of unrestricted tags, and a natural language description.

\textbf{\datalvis.}
While \data metadata contains a great deal of information about objects, Sketchfab's existing categorization scheme covers only 18 categories, too coarse for most applications. Object names, categories, and tags provide multiple potential categorizations at varying levels of specificity and with some inherent noise. However, for many existing computer vision tasks, it is useful to assign objects to a single category drawn from a predetermined set of the right size and level of semantic granularity.

We choose the categories from the LVIS dataset~\cite{gupta2019lvis} for categorizing a long-tail subset of objects in \data. We construct a 47K LVIS categorized object subset, called \datalvis, comprised of objects uniquely assigned to one of 1156 LVIS categories. We perform these assignments by first selecting 500 candidate objects per category using a combination of predictions from a CLIP classification model and candidates suggested by terms in their metadata. This combined pool contains objects visually resembling the target category (from the CLIP features of their thumbnail images) that might have missing metadata, as well as visually unusual instances of a category that are accurately named or tagged. These 250k candidate objects were then manually filtered and their assigned categories verified by crowdworkers. Since we only presented 500 object candidates per class, many popular categories, such as chair or car, have substantially more objects that could be included in \datalvis with future annotations.

\textbf{Animated objects and rigged characters.} \data includes 44K animated objects and over 63K objects self-categorized as characters. Examples of animations include fridge doors opening, animals running, and the hands on a clock moving. Rigged characters can be set up for animation and rendering, and may often come annotated with bone mappings. The vast scale of animations available in \data can support a wide range of research in temporal 3D learning, such as building text-based animation generative models~\cite{tevet2022human}, representing object changes over time with NERFs~\cite{pumarola2021d,park2021nerfies}, and temporal self-supervised learning via. future frame prediction~\cite{zellers2022merlot,jabri2020space}.

\begin{figure*}[b!]
    \includegraphics[width=\textwidth]{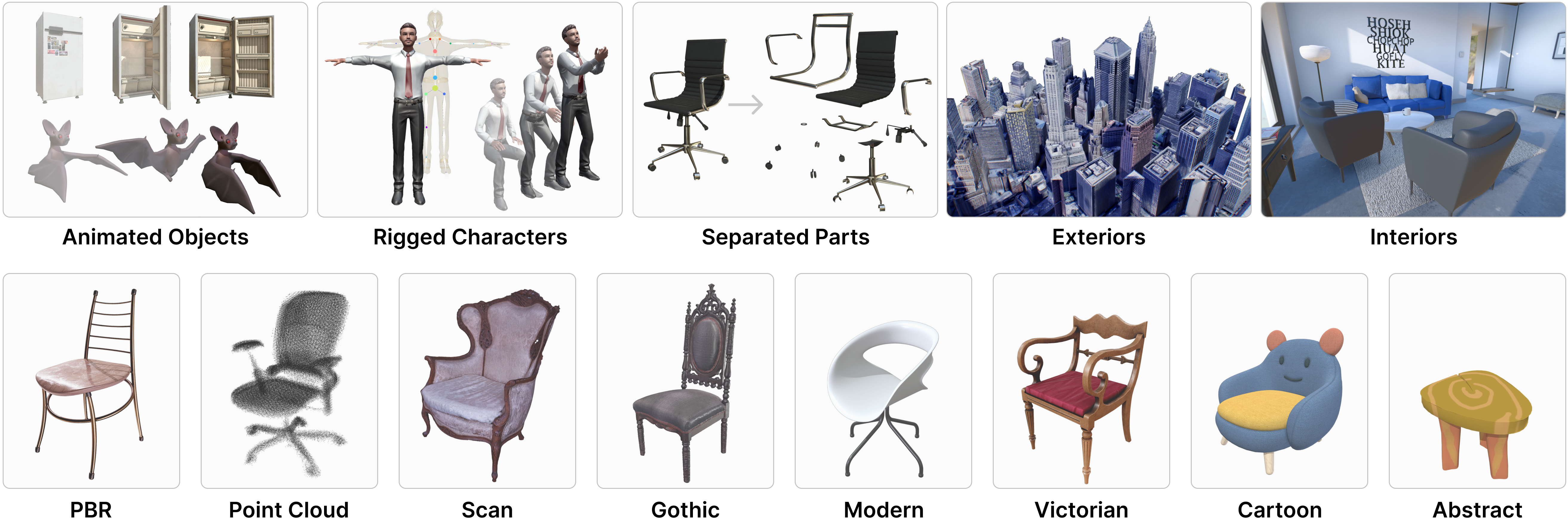}
    \caption{Highlights of the visual diversity of objects that appear in \data, including animated objects, rigged (body-part annotated) characters, models separatable into parts, exterior environments, interior environments, and a wide range visual styles.}
    \label{fig:visDiv}
    \vspace{-1em}
\end{figure*}

\textbf{Articulated objects.} Decomposing 3D objects into parts has led to a flurry of research in the past few years, including work in learning robotic grasping policies~\cite{xu2021adagrasp,xiang2020sapien}, 3D semantic segmentation~\cite{mo2019partnet}, and shape generation~\cite{mo2019structurenet}. Since many objects in \data were uploaded by artists, the objects often come separated into parts. Figure~\ref{fig:visDiv} shows an example, where a chair is separated by its backrest, wheels, and legs, among many smaller parts.

\textbf{Exteriors.} Photogrammetry and NERF advances have enabled the commercialization of capturing high-quality 3D objects of large exteriors by taking pictures ~\cite{tancik2022block,xiangli2022bungeenerf}. In \data, there are a large number of scanned buildings, cities, and stadiums. Figure~\ref{fig:visDiv} shows an example of a 3D object of NYC's skyline captured through a scan.

\textbf{\data-Interiors.} There are 16K+ interior scenes in \data, including houses, classrooms, and offices. The scenes often have multiple floors, many types of rooms, and are densely populated with objects from human input. Objects in the scenes are separable into parts, which allows them to be usable for interactive robotics, embodied AI, and scene synthesis. To put the scale of \data-Interiors in perspective, the number of scenes in \data-Interiors is significantly larger than the 400 or so existing hand-built interactive embodied AI scenes~\cite{kolve2017ai2thor,gan2020threedworld,li2021igibson,szot2021habitat}.

\textbf{Visual styles.} Objects in the world can be constructed in many styles and often differ in style based on the time-period, geographic location, and artist's style. \data objects cover a vast set of visual styles, including 3D scans, 3D modeled objects from virtually any platform, point clouds, and photo-realism via physically based rendering (PBR)~\cite{pharr2016physically}.
Moreover, instances of objects often appear with many styles, which is critical for training and evaluating robust computer vision models~\cite{radford2021learning}. Figure~\ref{fig:visDiv} shows examples of chairs in \data in many different styles, including Gothic, modern, Victorian, cartoon, and abstract.

\textbf{Statistics.} \data 1.0 includes 818K 3D objects, designed by 160K artists. There are ${>}$2.35M tags on the objects, with ${>}$170K of them being unique. We estimate that the objects have coverage for nearly 21K WordNet entities~\cite{miller1995wordnet} (see appendix for details). Objects were uploaded between 2012 and 2022, with over 200K objects uploaded uploaded just in 2021. Figure~\ref{fig:metadata vis} visualizes several statistics of the dataset, including the breakdown of objects into their self-assigned Sketchfab categories, a word cloud over the tags, a frequency plot of the tags, and the number of objects in \datalvis categories.

\section{Applications}
\label{sec:applications}

In this section, we present 4 initial distinct applications of \data, including 3D generative modeling, instance segmentation with CP3D, open-vocabulary ObjectNav, and analyze robustness in computer vision models. %

\subsection{3D Generative Modeling}\label{sec:3d-gen-modeling}

\begin{figure}[t!]
    \begin{subfigure}[t]{\columnwidth}
        \vspace{-0.225in}
        \makebox[\textwidth][c]{
        \includegraphics[width=1.075\textwidth]{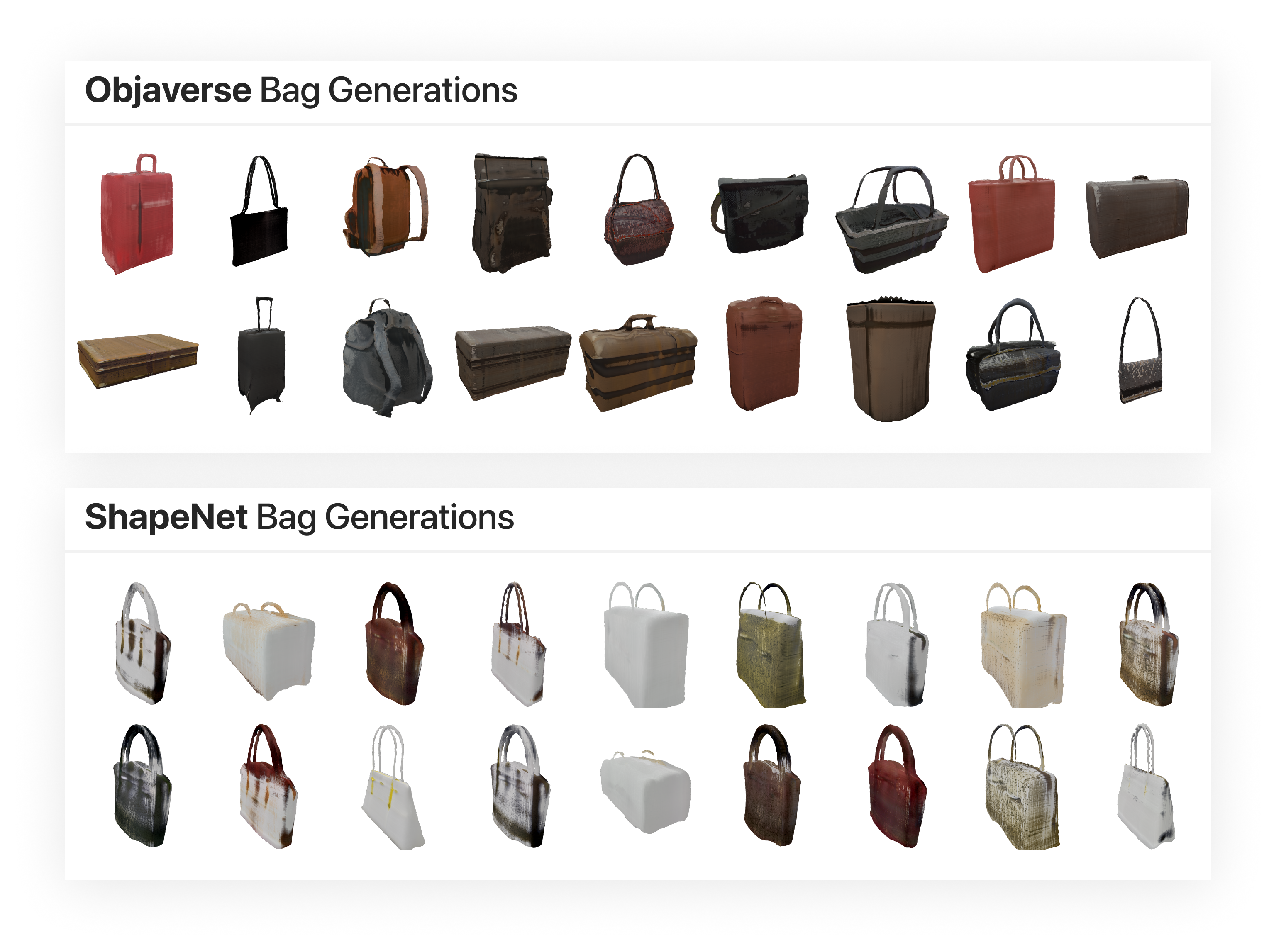}
        }
        \vspace{-0.31in}
        \caption{Comparison of \emph{Bag}s generated with \data and ShapeNet.}
    \end{subfigure}
    \begin{subfigure}[t]{\columnwidth}
        \vspace{-0.025in}
        \makebox[\textwidth][c]{
            \includegraphics[width=1.075\textwidth]{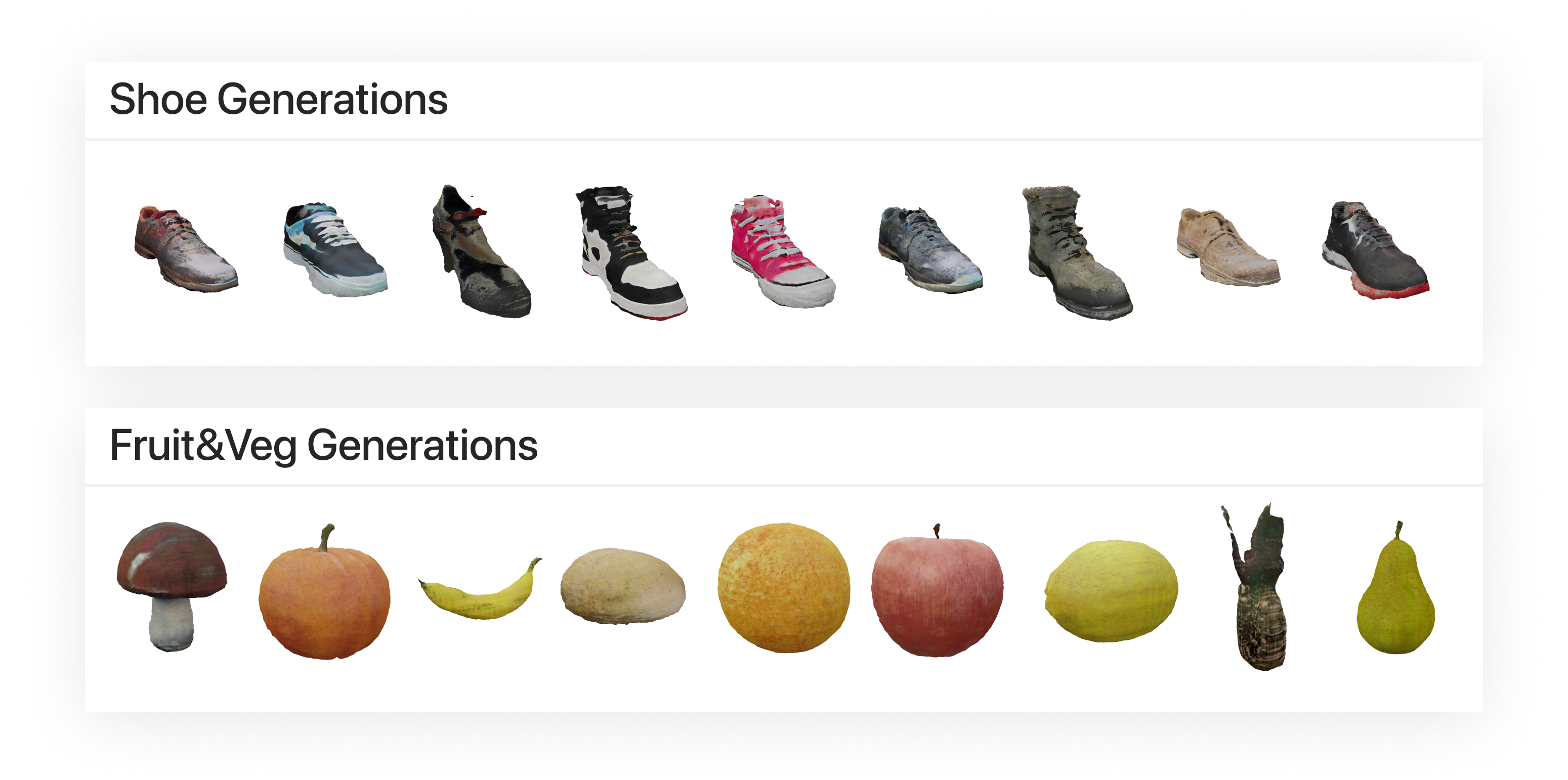}
        }
        \vspace{-0.31in}
        \caption{\emph{Shoe} and \emph{Fruit\&Veg.} generations.}
    \end{subfigure}
    \begin{subfigure}[t]{\columnwidth}
        \vspace{-0.025in}
        \makebox[\textwidth][c]{
            \includegraphics[width=1.075\textwidth]{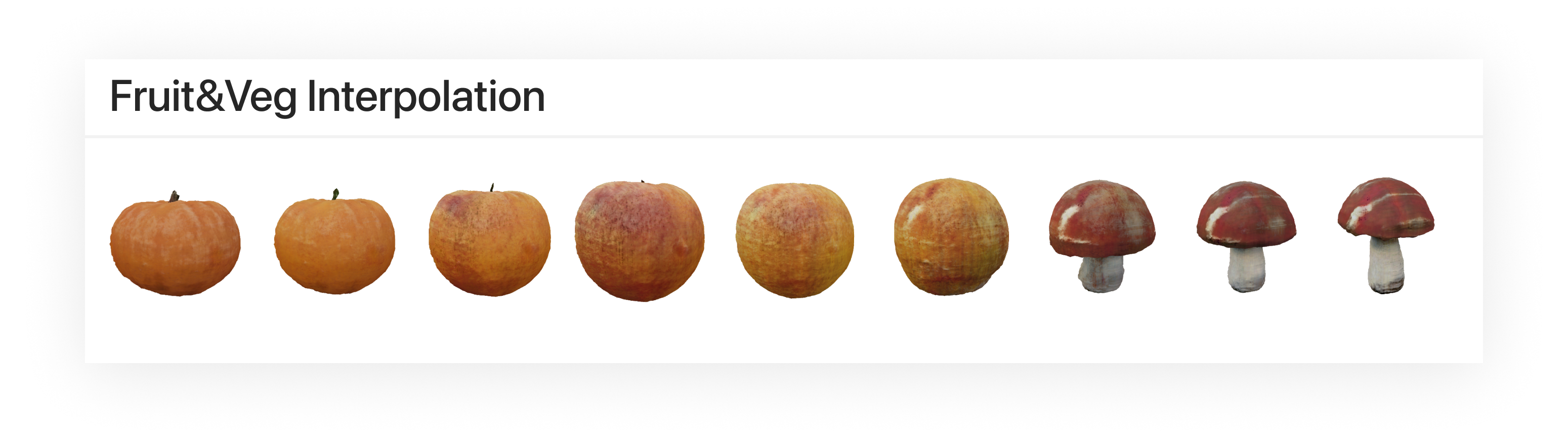}
        }
        \vspace{-0.31in}
        \caption{\emph{Fruit\&Veg.} interpolation.}
    \end{subfigure}
    \caption{(a) Example GET3D \emph{Bag} object generations using \data and ShapeNet models for training. (b) Additional \emph{Shoe} and \emph{Fruit\&Veg} generations from \data models. (c) models generated when interpolating between two, randomly sampled, latent encodings with our trained Fruit\&Veg{.} model; what appears to be a pumpkin smoothly transforms into a mushroom.}
    \vspace{-1em}
    \label{fig:text-to-3d}
\end{figure}

3D generative modeling has shown much improvement recently with models such as GET3D~\cite{gao2022get3d} delivering impressive high quality results with rich geometric details. GET3D is trained to generate 3D textured meshes for a category and produces impressive 3D objects for categories like \emph{Car}, \emph{Chair}, and \emph{Motorcycle} using data from ShapeNet~\cite{chang2015shapenet}. \data contains 3D models for many diverse categories including tail categories which are not represented in other datasets. It also contains diverse and realistic object instances per category. This scale and diversity can be used to train large vocabulary and high quality 3D generative models. In this work, we showcase the potential of this data as follows. We choose three categories of objects, \emph{Shoe}, \emph{Bag}, and \emph{Fruit\&Veg}, and subsample objects from \data to create three separate datasets containing, respectively, 143 shoes, 816 bags, and 571 fruits \& vegetables (116 apples, 112 gourds, 92 mushrooms, 68 bananas, 52 oranges, 52 pears, 31 potatoes 24 lemons, and 24 pineapples). For comparison, we also train a GET3D model on the set of 83 bags from the ShapeNet dataset. Fig.~\ref{fig:text-to-3d} shows a collection of 3D objects generated by our trained GET3D models. Qualitatively, the 3D-meshes generated by the \data-trained models are high-quality and diverse, especially when compared to the generations from the ShapeNet-trained model. To quantify this observation, we asked crowdworkers to rate the diversity of \emph{Bag} generations produced by the \data and ShapeNet trained models. When shown collections of nine randomly sampled generations from both models, workers rated the collection generated from the \data trained model as more diverse in appearance 91\% of the time.

Our fruits and vegetables, composed of 9 varieties produces perhaps the highest quality output, a promising signal that can inspire future work in text-to-3D generation.

\subsection{Instance Segmentation with CP3D}

\begin{figure}
    \vspace{-0.215in}
    \centering
    \makebox[\columnwidth][c]{
        \includegraphics[width=1.085\columnwidth]{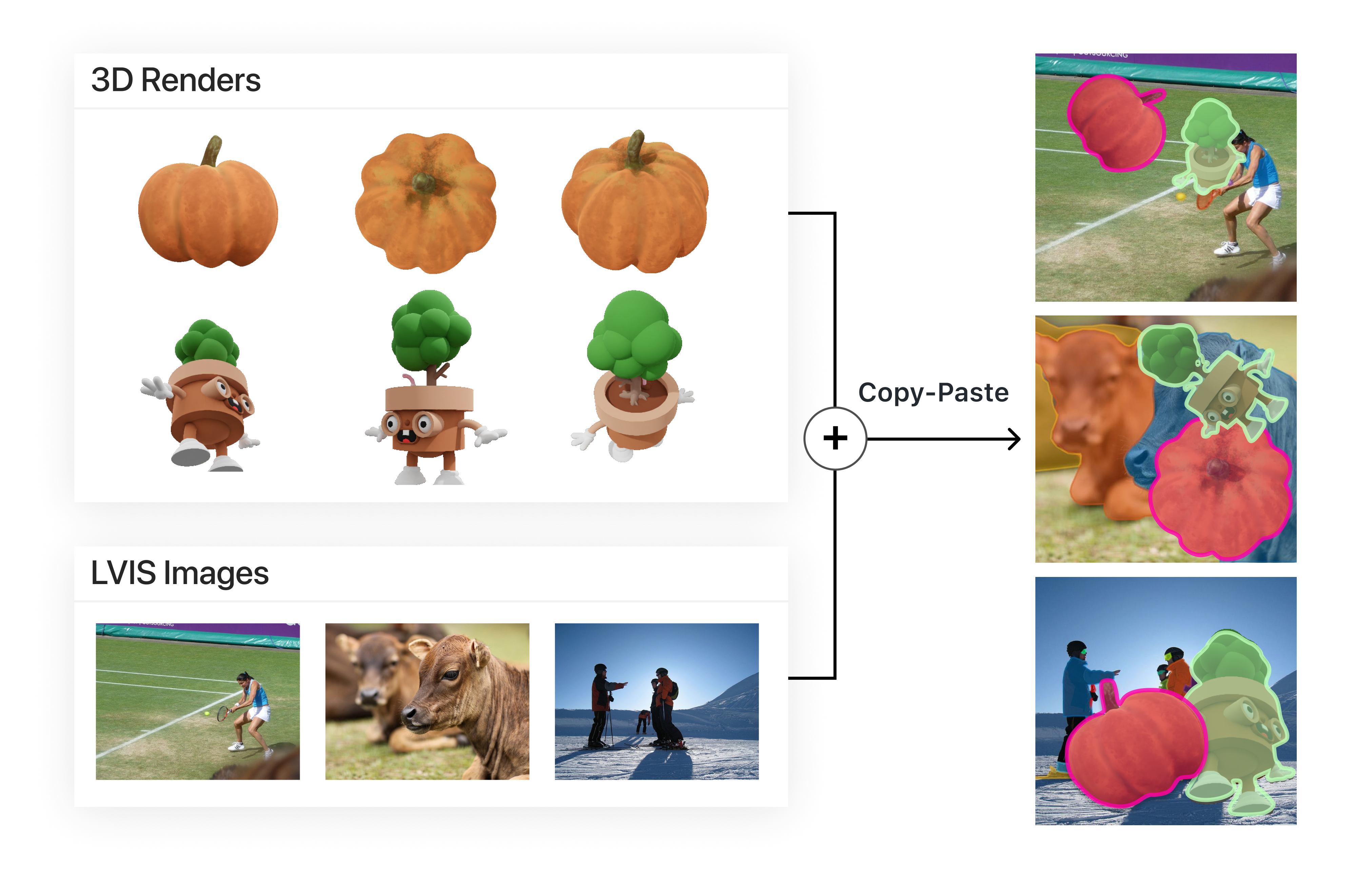}
    }
    \vspace{-0.31in}
    \caption{An illustration of 3DCP (3D copy-paste) for segmentation augmentation. We render 3D objects from multiple views and paste them over LVIS training images.}
    \label{fig:copy-paste-seg}
    \vspace{-1em}
\end{figure}

\begin{table}[b!]
\centering
\begin{tabular}{l ll ll ll ll}
\toprule
Method & AP & APr & APc & APf \\
\midrule
RFS~\cite{gupta2019lvis} & 23.7 & 13.3 & 23.0 & 29.0\\
EQLv2~\cite{tan2021equalization} & 25.5 & 17.7 & 24.3 & 30.2\\
LOCE~\cite{feng2021exploring} & 26.6 & 18.5 & 26.2 & 30.7\\
NorCal with RFS~\cite{pan2021model} & 25.2 & 19.3 & 24.2 & 29.0\\
Seesaw~\cite{wang2021seesaw} & 26.4 & 19.5 & 26.1 & 29.7\\
GOL~\cite{alexandridis2022long} & 27.7 & 21.4 & 27.7 & 30.4\\
GOL + 3DCP & \textbf{28.3} & \textbf{21.8} & \textbf{28.3} & \textbf{31.1} \\
\bottomrule
\end{tabular}
\caption{Comparison of our approach (GOL+3DCP) against SoTA Mask-RCNN ResNet-50 models on LVIS. We report results for APr, APc, and APf which measure AP for categories that are rare (appear in 1-10 images), common (appear in 11-100 images), and frequent (appear in $>$100 images), respectively}
\label{seg_table}
\vspace{-1em}
\end{table}

\begin{figure*}[t!]
    \centering
    \includegraphics[width=0.8\textwidth]{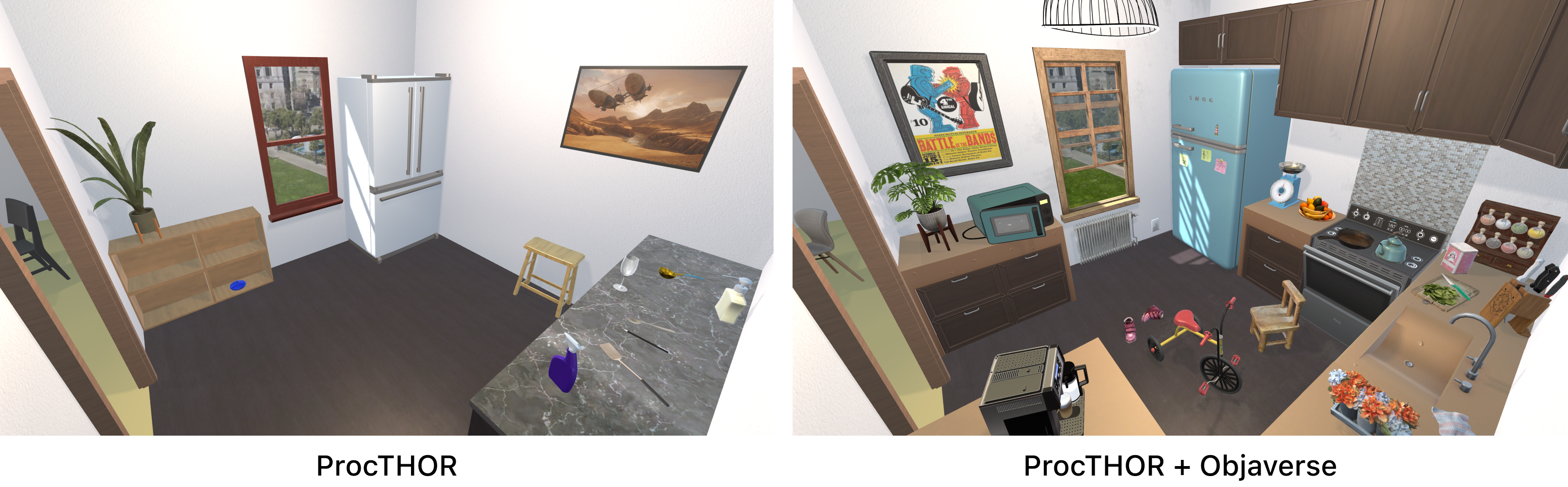}
    \vspace{-0.1in}
    \caption{An existing ProcTHOR scene (left) and a semantically similar ProcTHOR generatable scene with \data objects (right).}
    \vspace{-1.8em}
\end{figure*}

A key advantage of using simulated data for computer vision is that it is much cheaper to obtain expert annotations. Annotated \data objects can be rendered into images, allowing them to serve as a rich source of additional data that can be used to enhance model performance on 2D computer vision tasks. As a proof-of-concept demonstrating the effectiveness of this approach, we use segmented data from \data objects as auxiliary labels for training models on the LVIS dataset for Large Scale Instance Segmentation \cite{gupta2019lvis}. The LVIS dataset contains instance segmentation masks for 1200 object categories that occur throughout a set of 164k images. Recognition is especially challenging in this task due to the long tail of the object category distribution in this dataset. LVIS categories only contain an average 9 instances across the dataset, so training on simulated data is a promising approach for overcoming the challenges of learning in this low-sample regime. 

Using the LVIS-annotated subset of \data, we introduce 3DCP: an enhancement to the simple, but effective, copy-and-paste technique of\cite{ghiasi2021simple}. Figure~\ref{fig:copy-paste-seg} shows an example of the setup for 3DCP. Here, we render different views of 3D objects and paste them on-top of existing LVIS images. We render 5 distinct views of each object and cache them for use throughout training. During training, an image is selected for the copy-paste augmentation with 0.5 probability, and once selected, 1-3 images of randomly chosen LVIS-annotated \data objects are pasted onto the selected training image. The segmentation masks of the selected objects are added to the training image's annotation as well. Object images and masks are randomly scaled and translated before being pasted. We use this strategy to finetune the pretrained ResNet-50 Mask-RCNN \cite{he2017mask,he2016deep} of \cite{alexandridis2022long}. As shown in Tab.\ref{seg_table}, simply finetuning this model for 24 epochs yields performance gains across several metrics.

\vspace{-0.05in}

\subsection{Open-Vocabulary ObjectNav}

\vspace{-0.05in}

In this section, we introduce open-vocabulary ObjectNav, a new task propelled by the vast diversity of objects that appear in \data. Here, an agent is placed at a random starting location inside of a home and tasked to navigate to a target object provided from a text description (\eg \emph{``Raspberry Pi Pico''}). To facilitate this task, we procedurally generate 10K new homes in ProcTHOR~\cite{deitke2022procthor} fully populated with objects from \data-LVIS. Until now, ObjectNav tasks have focused on training agents to navigate to 20 or so target objects provided their category label~\cite{deitke2022retrospectives,ramrakhya2022habitat,deitke2022procthor}, and existing interactive embodied AI simulations, including ProcTHOR, only include around 2K total objects across around 100 object types~\cite{li2021igibson,deitke2022procthor,szot2021habitat}. In this work, we take a large step to massively scale the number of target objects used in ObjectNav (\textbf{20 $\to$ OpenVocab}), the number of objects available in simulation (\textbf{2K $\to$ 36K}), and the number of object types of the objects (\textbf{100 $\to$ 1.1K}).

\textbf{Object placement.} To make the placement of objects in the houses more natural, we use the \datalvis subset and annotate placement constraints for each object category. Specifically, we annotate if objects of a given category typically appears on the floor, on-top of a surface, or on a wall. If instances of the object category may appear on the floor, we also annotate whether it may appear in the middle of the scene (\eg a clutter object like a basketball) or on the edge of the scene (\eg a toilet or a fridge). For objects placed on the floor, we also to automatically detect flat regions on top of the object's mesh to place surface object types. The annotations are used by ProcTHOR for sampling objects to place in a scene. We also filter out \datalvis objects that do not appear inside of homes, such as a jet plane. Structural objects, like doors and windows, are inherited from ProcTHOR as they would require additional cleanup.

\textbf{Object size correction.} Objects in Sketchfab may be uploaded at unnatural scales (\eg a plant being as large as a tower). We therefore scale the objects to be of a reasonable size for them to look natural in a house. Here, for each object category, we annotate the maximum bounding box dimension length that every instance of the object category should be scaled to. For example, we annotate the maximum bounding box dimension for bookcase to be 2 meters 
and fork to be 0.18 meters. If a 3D modeled bookcase then has a bounding box of 20m$\times$6m$\times$3m, we shrink each side by a factor of $\max(20, 6, 3)/2 = 5$. 

\textbf{Preprocessing for AI2-THOR.} We add support to AI2-THOR for loading objects on the fly at runtime. Previously, all objects had to be stored in a Unity build, but such an approach is impractical when working with orders of magnitude more object data. For each object, we compress it with Blender~\cite{blender} by joining all of its meshes together, decimate the joined mesh such that it has at most 5K vertices, and bake all the UV texture maps into a single texture map. We then generate colliders using V-HACD~\cite{mamou2016volumetric} to support rigid-body interactions.

\textbf{Approach.}
Given procedural houses populated with \datalvis, the task is to navigate to the proximity of a chosen target object and invoke a task-completion action when the target object is in sight, given an open-vocabulary description formed with the template ``\texttt{a \{name\} \{category\}}". The \texttt{name} is the object name given by its creator, which is often descriptive. We filter each by whether it is detected as being written in English by a language detector \cite{joulin2016bag, joulin2016fasttext}, and fall back to a class-only description for non-English \texttt{name}. Examples of the possible expressions include \emph{``a victorian-monobike motorcycle"}, \emph{``a unicorn pony"}, or \emph{``a dino ghost lizard"}. The agent, similar to the ones in \cite{Khandelwal2022SimpleBE}, observes an RGB egocentric view of the environment, pre-processed by the visual branch of a frozen ResNet-50 CLIP model \cite{radford2021learning} -- the target description is pre-processed by the corresponding text branch. We train the agent with DD-PPO \cite{Wijmans2020DDPPOLN} and evaluate on houses with floor plans, objects, and descriptions unseen in training. We use the AllenAct~\cite{AllenAct} framework to train our agent. Our trained agent achieves a success rate of 19.9\%, for a random policy success of 5.1\%. For more details about the experiment refer to the appendix.%

\vspace{-0.05in}

\subsection{Analyzing Robustness}

\vspace{-0.05in}

\begin{figure}[t!]
    \centering
    \includegraphics[width=\columnwidth]{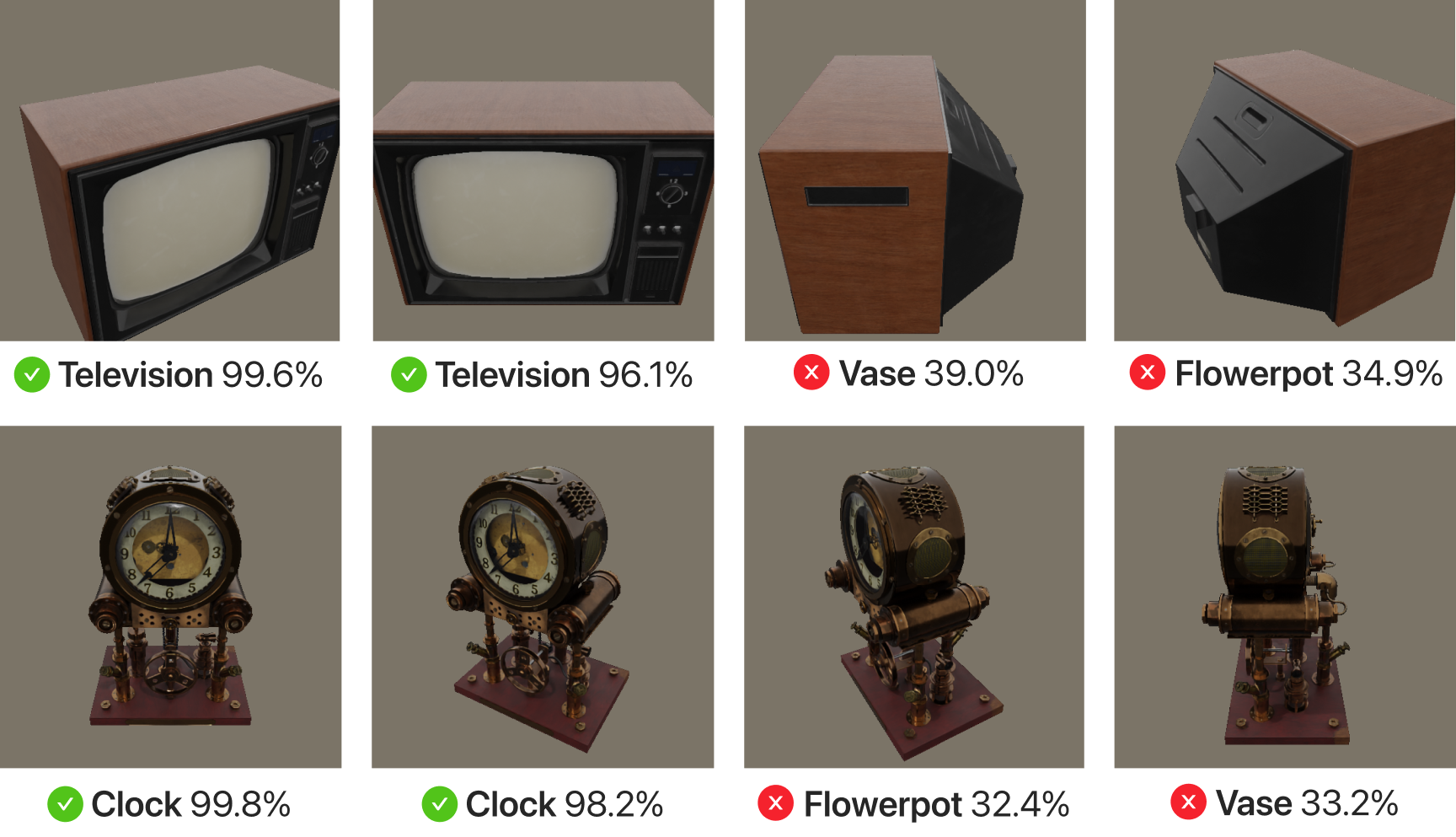}
    \caption{Examples of objects rendered from random orientations and their 0-shot classification categories with the CLIP ViT-B/32.}
    \label{fig:robustness}
    \vspace{-1em}
\end{figure}

A persistent bias present in many image datasets, \eg ImageNet~\cite{russakovsky2015imagenet}, is that the subjects of interest are generally photographed from a forward-facing, canonical, orientation. When, for example, taking a photograph of a television, few would choose to take this photograph crouched on the floor behind the television. This impact of this bias was studied by Alcorn~\etal~\cite{Alcorn2019StrikeAPose} who find that modern computer vision systems are highly susceptible to deviations from canonical poses. This is more than a theoretical problem: computer vision systems deployed in the real world will frequently encounter objects in non-canonical orientations and in many applications, \eg autonomous driving, it will be safety critical that they behave well.

Given the above, we adopt the experimental design of Alcorn~\etal and design, using \data assets, a benchmark for evaluating the robustness of state-of-the-art computer vision classification models to orientation shifts. In particular, for each object in our \datalvis subset, we render 12 images of the object from random orientations rendered upon a background with RGB values equalling the mean RGB values from ImageNet; see Fig.~\ref{fig:robustness} for examples. This ability to, at scale, render objects from random viewpoints is a practical impossibility in the real world but is made trivial when using 3D assets. We then evaluate several modern open-domain image-classification networks (constrained to the ${\approx}1{,}200$ \textsc{LVIS} categories) on these images and report 4 metrics for each model. These metrics include:

\begin{table}[t!]
    \centering
    \resizebox{\columnwidth}{!}{
    \begin{tabular}{l|cc|cc|cc}
        \toprule
        & \multicolumn{2}{c}{Random Rotation} & \multicolumn{2}{c}{Any Rotation}\\\cmidrule(lr){2-3}\cmidrule(lr){4-5}
        Model & Top-1 & Top-5 & Top-1 & Top-5 & $\Delta$Top-1 \\
        \midrule\\[-0.15in]
        OpenAI-400M~\cite{radford2021learning}\\
        RN50 & 21.4\% & 45.0\% & 43.9\% & 70.8\% & 22.5\% \\
        ViT-L/14 & 29.1\% & 54.5\% & \textbf{52.3\%} & 77.2\% & 23.2\%
        \\[0.05in]
        \midrule\\[-0.15in]
        LAION-400M~\cite{schuhmann2021laion}\\
        ViT-B/32 & 24.1\% & 48.5\% & 46.9\% & 74.2\% & 22.8\% \\
        ViT-L/14 & 30.6\% & 56.8\% & 50.5\% & 77.0\% & 19.9\% \\[0.05in]
        \midrule\\[-0.15in]
        LAION-2B~\cite{schuhmann2022laion}\\
        ViT-B/32 & 27.0\% & 51.8\% & 50.3\% & 76.1\% & 23.3\% \\
        ViT-L/14 & \textbf{32.9\%} & \textbf{59.2\%} & 52.1\% & \textbf{78.0\% } & 19.2\% \\
        ViT-H/14 & 32.3\% & 58.8\% & 50.1\% & 77.3\% & \textbf{17.8\%} \\[0.05in]
        \bottomrule
    \end{tabular}
    }
    \caption{Evaluating 0-shot CLIP classification models on our rotational robustness benchmark. $\Delta$Top-1 denotes the difference between \emph{Top-1 Any Rotation} and \emph{Top-1 Random Rotation}. Models are strongly overfit to standard views of objects.}
    \label{tab:robustness}
    \vspace{-1em}
\end{table}

\noindent$\bullet$ \emph{Top-1 Random Rotation} -- the frequency with which a model correctly classifies an image as belonging to the respective \textsc{LVIS} category.

\noindent$\bullet$ \emph{Top-1 Any Rotation} -- the frequency with which a model classifies an image correctly from at least one of the 12 random orientations.

This second metric is diagnostic and serves to represent a model's performance when shown an object from a canonical pose. We also have \emph{Top-5} variants of the above metric where the correct category need only be in the top 5 predictions from the model.
We report our results in Tab.~\ref{tab:robustness} in which we evaluate a variety of performant pretrained models. Comparing the gap in performance between the \emph{Top-$k$ Random Rotation} and \emph{Top-$k$ Any Rotation} metrics we find that model performance dramatically degrades when viewing objects from unusual orientations.

\vspace{-0.05in}

\section{Conclusion}
\label{sec:conclusion}
\vspace{-0.05in}

We present \data, a next-generation 3D asset library containing 818K high-quality, diverse, 3D models with paired text descriptions, titles, and tags. As a small glimpse of the potential uses of \data, we present four experimental studies showing how \data can be used to power (1) generative 3D models with clear future applications to text-to-3D generation, (2) improvements to classical computer vision tasks such as instance segmentation, (3) the creation of novel embodied AI tasks like Open Vocabulary Object Navigation, and (4) quantifying the rotational robustness of vision models on renderings of objects. We hope to see \data enable a new universe of new applications for computer vision.

{\small
\bibliographystyle{ieee_fullname}
\bibliography{egbib}
}

\newpage

\appendix

\input{supp.tex}

\end{document}

%% file: 01_intro_new.tex
Massive datasets have enabled and driven rapid progress in AI. Language corpora on the web led to large language models like GPT-3~\cite{brown2020language}; paired image and text datasets like Conceptual Captions~\cite{sharma2018conceptual} led to vision-and-language pretrained models like VilBERT~\cite{lu2019vilbert}; YouTube video datasets led to video capable models like Merlot-Reserve~\cite{zellers2022merlot}; and massive multimodal datasets like WebImageText~\cite{srinivasan2021wit} and LAION~\cite{schuhmann2021laion,schuhmann2022laion} led to models like CLIP~\cite{radford2021learning} and StableDiffusion~\cite{rombach2022high}. These leaps in dataset scale and diversity were triggered by moving from manually curated datasets to harnessing the power of the web and its creative content.

In contrast to the datasets described above, the size of the datasets we are feeding to our, data-hungry, deep learning models in many other areas of research is simply not comparable. For instance, the number of 3D assets used in training generative 3D models is, maximally, on the order of thousands~\cite{gao2022get3d} and the simulators used to train embodied AI models typically have only between a few dozen to a thousand unique scenes~\cite{kolve2017ai2thor,szot2021habitat,ramrakhya2022habitat,li2021igibson}. The startling advances brought about by developing large-scale datasets for images, videos, and natural language, demand that an equivalent dataset be built for 3D assets.

We present \data 1.0, a large scale corpus of high-quality, richly annotated, 3D objects; see Fig.~\ref{fig:teaser}. Objects in our dataset are free to use\footnote{Creative Commons license} and sourced from Sketchfab, a leading online platform for managing, viewing, and distributing 3D models. In total, \data contains over \textbf{800K} 3D assets designed by over \textbf{100K} artists which makes this data large and diversely sourced.
Assets not only belong to varied categories like animals, humans, and vehicles, but also include interiors and exteriors of large spaces that can be used, \eg, to train embodied agents. \data is a universe of rich 3D data with detailed metadata that can support many different annotations to enable new applications. 
With this remarkable increase in scale, we see an incredible opportunity for \data to impact research progress across domains. In this work, we provide promising results to answer three questions.

\textbf{Can 3D vision benefit from a large-scale dataset?}
First, as a 3D asset resource, \data can support the exciting field of 3D generative modeling. We use data extracted from \data to train generative models for single and multiple categories using GET3D~\cite{gao2022get3d} and find that we are able to generate high-quality objects and, moreover, that our generated objects are found by human annotators to be more diverse than those generated by a model trained on ShapeNet objects in 91\% of cases.

\textbf{Can the diversity of 3D models help improve classical 2D vision task performance?}
To answer this question, we use the diversity of \data
to improve the performance of long tail instance segmentation models. Instance segmentation data can be expensive to obtain owing to the cost of annotating contours around objects. The recent LVIS dataset contains annotations for 1{,}230 categories but the task remains very challenging for present day models, particularly on tail categories that have few examples. We show that increasing the volume of data by leveraging a simple Copy{+}Paste augmentation method with \data assets can improve the performance of state-of-the-art segmentation methods.

We also use \data to build a benchmark for evaluating the robustness of state-of-the-art visual classification models to perspective shifts. We render objects in \data from random orientations, which is how one might expect to see them in the real world and test the ability of CLIP-style visual backbones to correctly classify these images. Our experiments show that current state-of-the-art models' performance degrades dramatically in this setting when viewing objects from arbitrary views.
\data allows us to build benchmarks to test (and potentially train) for orientation robustness for a long tail distribution of asset categories. Building such benchmarks is made uniquely possible by the scale and diversity of 3D assets in \data. This would simply not be feasible to create in the real world nor can they be generated from existing 2D images.

\begin{figure*}[t]
    \begin{subfigure}[t]{0.725\textwidth}
    \centering
    \includegraphics[width=\textwidth]{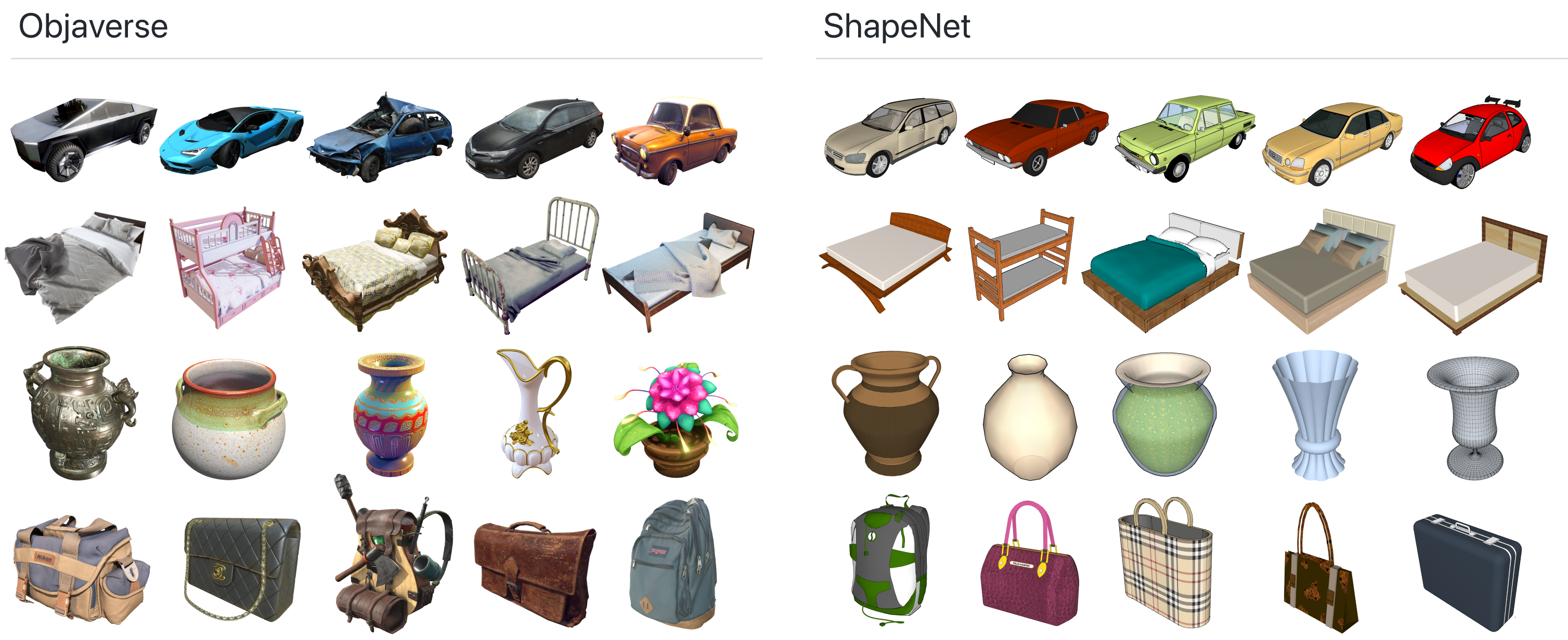}
    \end{subfigure}
    \hfill
    \begin{subfigure}[t]{0.25\textwidth}
        \vspace{-1.85in}
        \centering
        \resizebox{\textwidth}{!}{%
        \begin{tabular}{lccccccc}
            \toprule
            Dataset & \# Objects & \# Classes \\
            \midrule
            YCB~\cite{calli2015benchmarking} & 77 & 5 \\[\s]
            BigBIRD~\cite{Singh2014BigBIRDAL} & 125 & -- \\[\s]
            KIT~\cite{kasper2012kit} & 219 & 145 \\[\s]
            IKEA~\cite{lim2013parsing} & 219 & 11 \\[\s]
            Pix3D~\cite{sun2018pix3d} & 395 & 9 \\[\s]
            GSO~\cite{downs2022google} & 1K & 17 \\[\s]
            EGAD~\cite{morrison2020egad} & 2K & -- \\[\s]
            PhotoShape~\cite{park2018photoshape} & 5K & 1 \\[\s]
            ABO~\cite{collins2022abo} & 8K & 63 \\[\s]
            3D-Future~\cite{fu20213d} & 10K & 34 \\[\s]
            ShapeNet~\cite{chang2015shapenet} & 51K & 55 \\[\s]
            \textbf{Objaverse} & \textbf{818K} & \textbf{21K} \\
            \bottomrule
        \end{tabular}
        }
    \end{subfigure}
    \caption{Comparison between \data and existing 3D object datasets. (\textit{Left:}) Visual comparison of instances from \data and ShapeNet for the categories of \textsc{Car}, \textsc{Bed}, \textsc{Vase}, and \textsc{Bag}. \data instances are substantially more diverse since objects can come from many 3D content creation platforms, whereas ShapeNet models look more similar and all come from SketchUp, a 3D modeling platform built for simple architectural modeling. (\textit{Right:}) Scale comparison table between existing 3D object datasets.}
    \label{fig:shapenet-comparison}
    \vspace{-1em}
\end{figure*}

\textbf{Can a large-scale 3D dataset help us train embodied agents performant embodied agents?
}
We use assets in \data to populate procedurally generated simulated environments in ProcTHOR~\cite{deitke2022procthor} that are used to train Embodied AI agents.
This results in an orders of magnitude increase in the number of unique assets available for use in ProcTHOR scenes (previously limited to AI2-THOR's~\cite{kolve2017ai2thor} asset library of a few thousand unique instances each assigned to one of 108 object categories).
Using \data populated scenes enables open vocabulary object navigation from any text description. In this paper, we provide quantitative results for navigating to 1.1K semantic object categories, roughly a 50x increase.  %

These findings represent just a small fraction of what can be accomplished using \data. We are excited to see how the research community will leverage \data to enable fast and exciting progress in 2D and 3D computer vision applications and beyond.

%% file: supp.tex
\appendix

\section{Instance Segmentation with CP3D}

\paragraph{Model.} We use the Mask-RCNN\cite{he2017mask} model of \cite{alexandridis2022long} with a ResNet-50 backbone\cite{he2016deep}; no additional changes to their model are made. Instead of a softmax activation, the model uses a Gumbel activation, given by the formula $\eta(q) = \exp(-\exp(-q))$, to transform logits into probabilities. More details about the model and activation can be found in \cite{alexandridis2022long}.

\paragraph{Training.} We take the pretrained ResNet-50 Mask-RCNN checkpoint of \cite{alexandridis2022long} and finetune the model for 24 epochs with the CP3D augmentation integrated into the training pipeline. We use a batch size of 64 and a learning rate of $0.002$.

\paragraph{Additional Results} Here we report detection metrics in addition to the segmentation results reported in the paper in Table \ref{seg_table}. Notably, we see an impressive gain of two points on AP for rare categories.

\begin{table}[h!]
\centering
\begin{tabular}{l ll ll ll ll}
\toprule
Method & AP & APr & APc & APf \\
\midrule
GOL~\cite{alexandridis2022long} & 27.5 & 19.8 & 27.2 & 31.2\\
GOL + 3DCP & \textbf{28.9} & \textbf{21.8} & \textbf{28.7} & \textbf{32.2} \\
\bottomrule
\end{tabular}
\caption{\textbf{Detection results for bounding box AP category metrics.} APr, APc, and APf measure AP for categories that are rare (appear in 1-10 images), common (appear in 11-100 images), and frequent (appear in $>$100 images), respectively.}
\label{seg_table_supp}
\vspace{-1em}
\end{table}

\section{Open-Vocabulary ObjectNav}

\begin{figure}
    \centering
    \includegraphics[width=0.9\linewidth]{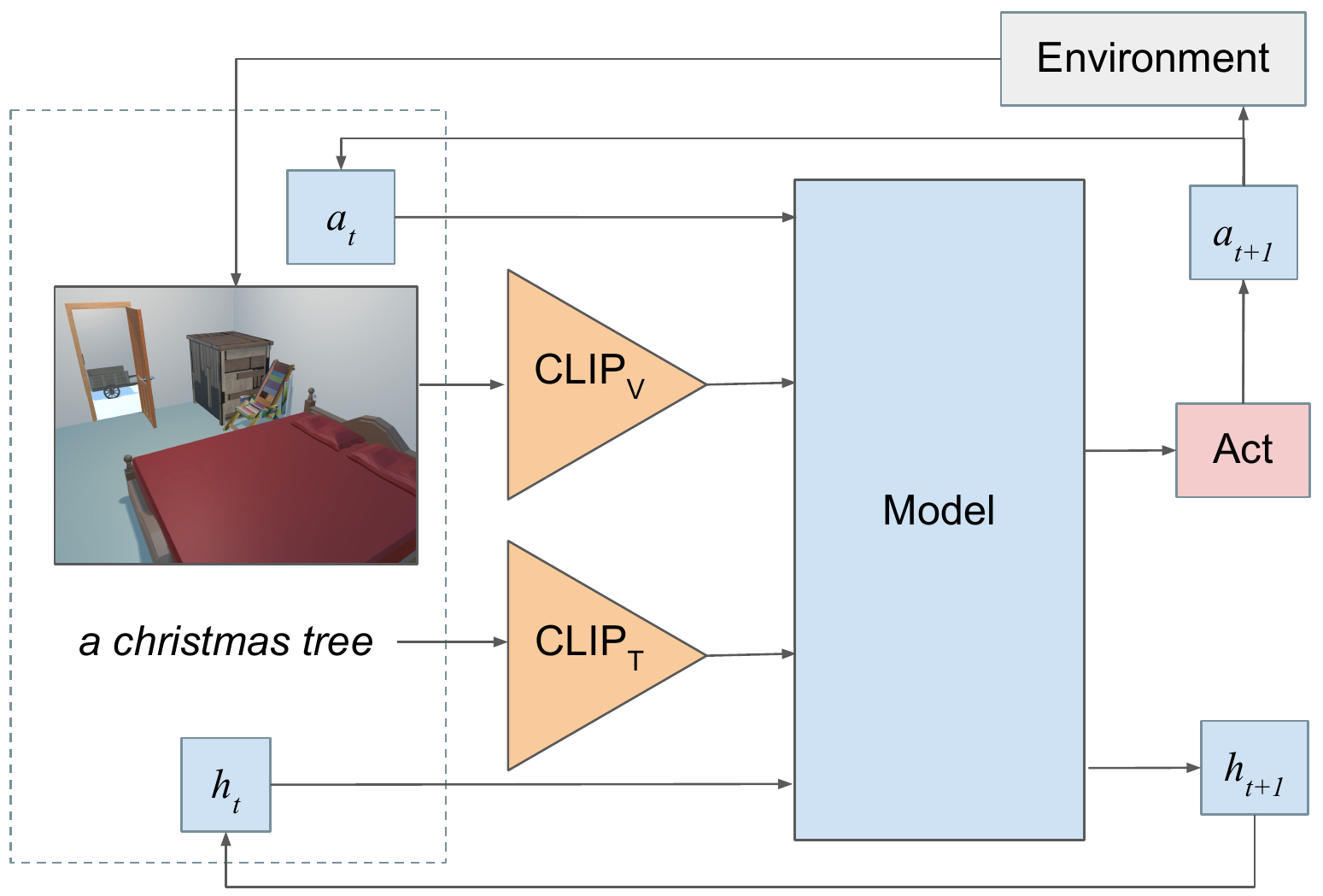}
    \caption{\label{fig:openvocabmodel}\textbf{Open-Vocabulary ObjNav Model overview.} The ObjectNav model (employing an RNN) uses the high-level architecture illustrated here, where it receives features from the visual and target object description encoders, besides previous hidden units and actions as input, and outputs the next action.}
\end{figure}

\paragraph{Model.} The agent's embodiment is a simulated LoCoBot \cite{locobot}. The action space consists of six actions: \textsc{MoveAhead}, \textsc{RotateLeft}, \textsc{RotateRight}, \textsc{End}, \textsc{LookUp}, and \textsc{LookDown}. Given the excellent exploration capabilities of EmbCLIP \cite{Khandelwal2022SimpleBE, deitke2022procthor}, we opt to keep the same overall architecture, just replacing the learned embedding for target types in prior work by a linear projection of the text branch output of CLIP for the target description, as shown in Fig.~\ref{fig:openvocabmodel}. Additionally, in order to provide more information about the target and the current visual input, we increase the respective internal representations for each modality from the original 32-D to 256-D. Note that our model does not employ the alternative zero-shot design described in \cite{Khandelwal2022SimpleBE}, where the target description is not observed by the agent's RNN. Given the scale of \datalvis, we can train agents with good generalization following a more standard design.

\paragraph{Training.} For training, we use ProcTHOR to procedurally generate 10,080 houses. Each house has up to three rooms, entirely populated with \datalvis assets except for structural components like doors and windows, which are inherited from ProcTHOR \cite{deitke2022procthor}. We sample targets corresponding to LVIS categories for which a single instance is present in the scene, resulting in a total of 9,421 unique assets corresponding to 262 categories targeted during training. Training uses DD-PPO \cite{Wijmans2020DDPPOLN} and is distributed across 28 GPUs on 7 AWS g4dn.12xlarge machines, with each 
 GPU hosting 360 houses and the subset of \datalvis assets populating them. The training hyperparameters, identical to the ones in \cite{deitke2022procthor}, and the 262 training target categories are listed in Table~\ref{tab:hyperparams} and Table~\ref{tab:all_obj_types}, respectively.

\begin{table}
    \centering
    \begin{tabular}{ll}
        \toprule
        \textbf{Hyperparameter}\qquad\qquad\qquad & \textbf{Value}\\
        \midrule
        Discount factor ($\gamma$) & 0.99 \\
        GAE parameter ($\lambda$) & 0.95 \\
        Value loss coefficient & 0.5 \\
        Entropy loss coefficient & 0.01 \\
        Clip parameter ($\epsilon$) & 0.1 \\
        Rollout horizons & 32, 64, 128 \\
        Rollout timesteps & 20 \\
        Rollouts per minibatch & 1\\
        Learning rate & $3\cdot10^{-4}$\\
        Optimizer & Adam \cite{kingma2014adam} \\
        Gradient clip norm & 0.5 \\
        \toprule\\[-0.05in]
    \end{tabular}
    \caption{\textbf{Training hyperparameters for Open-Vocabulary ObjectNav.}}
    \label{tab:hyperparams}
\end{table}

\begin{table*}
\centering
\emph{
\begin{tabular}{l}
\toprule
Bible, Christmas tree, Rollerblade, alligator, ambulance, amplifier, arctic (type of shoe), armor,\\banner, barbell, barrel, barrow, baseball bat, basketball, bat (animal), bath mat, beachball, bear, bed,\\beetle, bench, beret, bicycle, binder, binoculars, bird, blackberry, bookcase, boot, bottle, bowling ball,\\bullhorn, bunk bed, bus (vehicle), butterfly, cab (taxi), cabinet, canoe, cape, car (automobile), card,\\cardigan, carnation, cart, cassette, cat, chair, chaise longue, chicken (animal), clothes hamper, coatrack,\\coffee table, cone, convertible (automobile), cornice, cow, cowboy hat, crab (animal), crate, crossbar,\\cube, cylinder, deck chair, deer, desk, dinghy, dirt bike, dog, dollhouse, doormat, dove, drawer, dresser,\\duckling, dumbbell, dumpster, easel, elephant, elk, fan, ferret, file cabinet, fireplace, fireplug,\\fishing rod, flag, flagpole, flamingo, flip-flop (sandal), flipper (footwear), foal, football (American),\\footstool, forklift, frog, futon, garbage, gargoyle, giant panda, giraffe, golf club, golfcart, gondola (boat),\\goose, gorilla, gravestone, grill, grizzly, grocery bag, guitar, handcart, hat, heater, hockey stick, hog,\\horse, horse carriage, jeep, kayak, keg, kennel, kitchen table, kitten, knee pad, ladder, ladybug,\\lamb (animal), lamp, lamppost, lawn mower, legging (clothing), lion, lizard, locker, log, loveseat,\\machine gun, mailbox (at home), manhole, mascot, mast, milk can, minivan, monkey, mop, motor,\\motor scooter, motor vehicle, motorcycle, mushroom, music stool, nut, ostrich, owl, pajamas,\\parasail (sports), parka, penguin, person, pet, pew (church bench), piano, pickup truck, pinecone,\\ping-pong ball, playpen, pole, polo shirt, pony, pool table, power shovel, propeller, pug-dog, pumpkin,\\rabbit, radiator, raincoat, ram (animal), rat, recliner, refrigerator, rhinoceros, rifle, road map,\\rocking chair, router (computer equipment), runner (carpet), saddle (on an animal), saddle blanket,\\saddlebag, sandal (type of shoe), scarecrow, scarf, sculpture, seabird, shark, shepherd dog, shield, shirt,\\shoe, sink, skateboard, ski parka, skullcap, snake, snowmobile, soccer ball, sock, sofa, sofa bed,\\solar array, sparkler (fireworks), speaker (stero equipment), spear, spider, sportswear, statue (sculpture),\\step stool, stepladder, stool, subwoofer, sugarcane (plant), suit (clothing), suitcase, sunhat, surfboard,\\sweat pants, sweater, swimsuit, table, tape measure, tarp, telephone pole, television camera, tennis ball,\\tennis racket, tights (clothing), toolbox, tote bag, towel, trailer truck, trampoline, trash can, tricycle,\\trousers, truck, trunk, turtle, tux, underdrawers, vacuum cleaner, vending machine, vest, wagon wheel,\\water ski, watering can, wet suit, wheel, window box (for plants), wok, wolf, \emph{and} wooden leg.\\
\bottomrule
\end{tabular}%
}
\caption{\label{tab:all_obj_types}\textbf{Training target types for Open-Vocabulary ObjectNav.}}
\end{table*}

\paragraph{Testing.} For testing, we sample 150 episodes for each of 30 target categories, which are a subset of the training target categories. The resulting 4,500 episodes are sampled from 151 procedural houses not seen during training. The 30 testing target categories are listed in Table~\ref{tab:test_obj_types}. For the results provided in the main paper, the agent is trained for just 18 million simulation steps, but the resulting policy already shows reasonable performance given the variety of targets and scenes. Improved performance can be achieved with extended training (e.g., after approx.~460 million steps, the success rate is 33.0\%). 

\begin{table*}
\centering
\emph{
\begin{tabular}{l}
\toprule
Christmas tree, bed, bench, blackberry, chair, chicken (animal), dog, easel, elk, fireplug, forklift, garbage,\\gargoyle, guitar, mascot, motor, penguin, pony, pool table, radiator, rifle, scarf, sock,\\speaker (stero equipment), sportswear, sweat pants, trash can, trunk, wet suit, \emph{and} wheel.\\
\bottomrule
\end{tabular}%
}
\caption{\label{tab:test_obj_types}\textbf{Testing target types for Open-Vocabulary ObjectNav.}}
\end{table*}

\section{Composition}

\paragraph{Human subjects data.} A portion of the data included in \data is generated by human subjects (\ie crowdworkers recruited through Amazon's Mechanical Turk platform) as outlined in Section 3 and detailed below. The collection process has been reviewed and approved for release by an Institutional Review Board.

\begin{figure}[tp]
     \centering
     \begin{subfigure}[b]{\columnwidth}
         \centering
         \includegraphics[width=\textwidth]{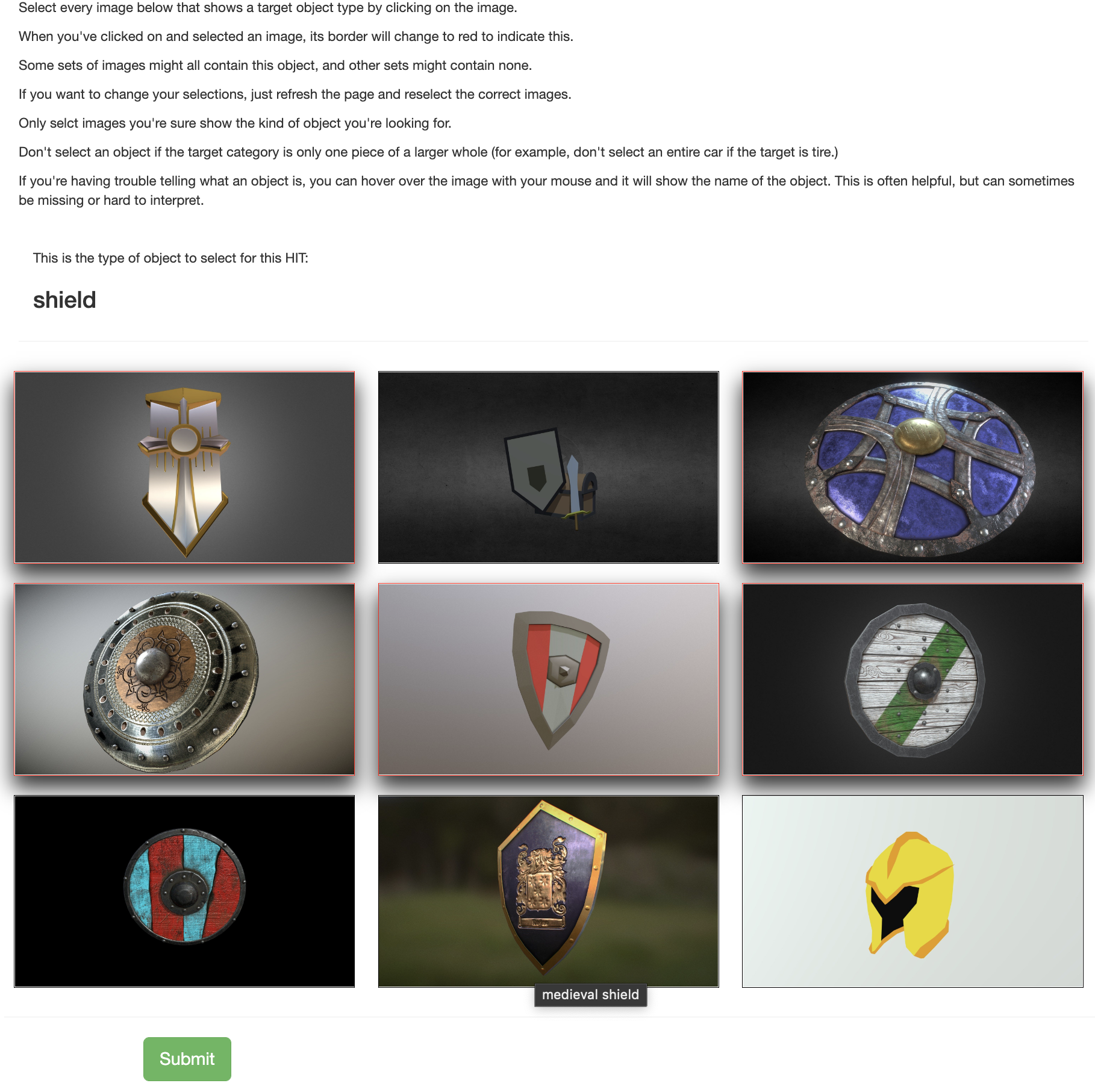}
         \caption{Screenshot of \datalvis categorization task.}
         \label{figures:category_annotation} 
     \end{subfigure}
     \hfill
     \vspace{0.25cm}
     \begin{subfigure}[b]{\columnwidth}
         \centering
         \includegraphics[width=\textwidth]{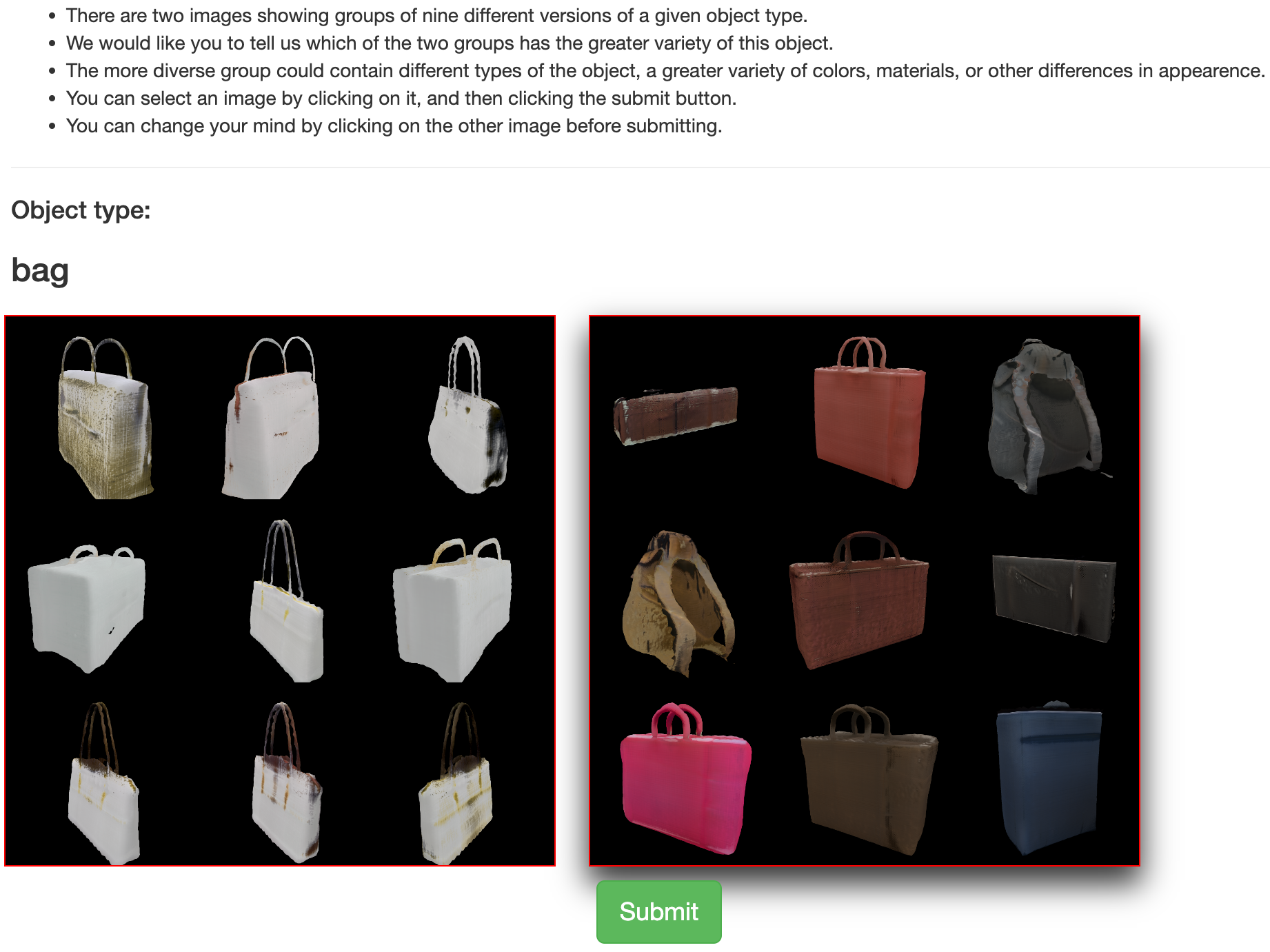}
         \caption{Screenshot of relative diversity rating task.}
         \label{figures:div_annotation} 
     \end{subfigure}
     \caption{\textbf{Data collection interfaces.}}
     \label{fig:interfaces}
\end{figure}

\paragraph{Data collection interfaces.} Human annotators were used to provide the category labels for \datalvis as described in Section 3. This task was accomplished by first creating sets of 500 candidate objects for each LVIS category. These candidate sets included objects visually resembling the target category (as ranked by the CLIP features of their thumbnail images), as well as instances whose metadata contained terms with a high similarity to the target category (as ranked by their GloVe vector similarity~\cite{pennington2014glove}). Candidate objects were shown to crowdworkers nine at a time, and they were asked to mark objects that were members of the category, as shown in Figure~\ref{fig:interfaces} a. In addition to the visual reference for each object, annotators also had access to the object's name and were encouraged to use this when helpful. Human annotators were also used to rate the relative diversity of  of 3D objects generated by models trained using \data and ShapeNet. The user interface and instructions for this task are shown in Figure~\ref{fig:interfaces} b. Two sets of nine objects generated by each model were shown with random left-right orientations, and workers were asked to choose the set exhibiting the greater variety in appearance.

\section{Estimating Coverage}

We use OpenAI's CLIP ViT-B/32 model to estimate the categorical coverage of the objects in \data. Specifically, for each object, we compute the CLIP image embedding from the thumbnail and the cosine similarity between an text embedding of each WordNet entity~\cite{fellbaum2010wordnet}. The entity with the maximum cosine similarity is then assigned as the object's entity. The WordNet entities are textually encoded in the form, ``a \{entity\} is a \{definition\}'', which is loosely inspired by CuPL~\cite{pratt2022does}. For instance, we might have ``a \emph{bat} is a nocturnal mouselike mammal with forelimbs modified to form membranous wings and anatomical adaptations for echolocation by which they navigate'' or ``a \emph{bat} is a club used for hitting a ball in various games''. Computing the nearest WordNet entity for each object gave us an estimated coverage of 20.8K entities.